# Improving Convolutional Neural Networks Via Conservative Field Regularisation and Integration

Dominique Beaini[a], Sofiane Achiche[a], Maxime Raison[a]

[a]Polytechnique Montreal, 2900 Edouard Montpetit Blvd, Montreal, H3T 1J4, Canada

## Abstract

Current research in convolutional neural networks (CNN) focuses mainly on changing the architecture of the networks, optimizing the hyper-parameters and improving the gradient descent. However, most work use only 3 standard families of operations inside the CNN, the convolution, the activation function, and the pooling. In this work, we propose a new family of operations based on the Green's function of the Laplacian, which allows the network to solve the Laplacian, to integrate any vector field and to regularize the field by forcing it to be conservative. Hence, the Green's function (GF) is the first operation that regularizes the 2D or 3D feature space by forcing it to be conservative and physically interpretable, instead of regularizing the norm of the weights. Our results show that such regularization allows the network to learn faster, to have smoother training curves and to better generalize, without any additional parameter. The current manuscript presents early results, more work is required to benchmark the proposed method.

## 1 Introduction

Since the year 2015, the convolutional neural networks (CNN) rose quickly to become the best machine learning technique used to solve many computer vision problems such as classification [1,2], edge detection [3,4], skeleton extraction [5] and salient object detection [6,7]. In fact, recent algorithms perform near the human-level [3]. For each of these tasks, major work is done on optimizing the architecture and the hyperparameters of the networks, with some work done on optimizing the gradient descent [8]. However, most networks only use 3 families of operations: the convolutional kernel (with trainable weights), the pooling and the activation function [2,3,5,9].

**The objective of this study** is to present a new family of operation based on the Green's function convolution (GF), which allows solving the Laplacian, to integrate any vector field and to regularize it by making it conservative. These new operations have many advantages since they allow to propagate information in the whole feature space and are not limited by the receptive field of the convolutional layer. Furthermore, by enforcing conservative fields, they regularize the network into having a physically interpretable feature space, which we showed to improve the training time, to reduce the noise in the training curve, and to better generalize to more complex images.

The Green's function was previously used in computer vision mainly for image editing in the gradient domain [10,11], but it was only recently used in our previous work for CNNs [12] for the task of salient object detection. In the current work, we generalize the idea that was previously proposed and show that it can be applied to any kind of 2D or 3D CNN.

## 2 Methodology

### 2.1 The numerical Green's function

This subsection shows how to build the numerical Green's function using the convolution theorem and the numerical Fourier transform. It summarizes the main results from our previous work [11].

To generate the Green's function, we first define the zero-padded matrices $\breve{K}_{\nabla^2}$ and $\breve{\delta}$ in equations (1) and (2), where the top left corner are the $3 \times 3$ Laplacian and Dirac kernels and the rest of the matrices are 0-valued. Both matrices are the same size as the Laplacian matrix $L$ that we want to solve.

$$\breve{K}_{\nabla^2} \equiv \begin{bmatrix} 0 & -1 & 0 & \cdots & 0 \\ -1 & 4 & -1 & & \\ 0 & -1 & 0 & & \\ \vdots & & & \ddots & \\ 0 & & & & 0 \end{bmatrix}_{size(L)} \tag{1}$$

$$\breve{\delta} \equiv \begin{bmatrix} 0 & 0 & 0 & \cdots & 0 \\ 0 & 1 & 0 & & \\ 0 & 0 & 0 & & \\ \vdots & & & \ddots & \\ 0 & & & & 0 \end{bmatrix}_{size(L)} \tag{2}$$

Using these matrices, we find the Green's function in the Fourier domain $\breve{V}_{mono}^{\mathcal{F}}$ in equation (3), where $\mathcal{F}$ is the numerical Fourier transform.

$$\check{V}^{\mathcal{F}}_{mono} = \frac{\mathcal{F}(\check{\delta})}{\mathcal{F}(\check{K}_{\nabla^2})} \qquad (3)$$

Hence, we can solve the Laplacian $L$ and find it's associated potential $I_c$ using equation (4), where $\mathcal{R}$ is the real part of a complex number and ∘ is the Hadamard element-wise product. We note that the integration constant $c$ can be defined arbitrarily, depending on the application.

$$I_c = \mathcal{R}\left(\mathcal{F}^{-1}\big(\mathcal{F}(L) \circ \check{V}^{\mathcal{F}}_{mono}\big)\right) + c \qquad (4)$$

### 2.2 Pseudo-code

In this section, we provide some Python-based pseudo-codes, based on our previous work [11].

First, Algorithm A shows how to compute the Green's function `green_F` (previously noted $\check{V}^{\mathcal{F}}_{mono}$) for an image of size `image_size`. Then, Algorithm B shows how the previous Green's function is used to solve the given Laplacian `padded_L` with equation (4).

For both algorithms, we must keep in mind that the 2D FFT `fft` and inverse FFT `ifft` produce complex outputs, meaning that the products and division must be used accordingly. The code simplicity allowed us to implement the solver using Matlab, C++ (OpenCV) and Python (Tensorflow and Pytorch).

For a 3D CNN, the procedure is identical, except that the Dirac and Laplacian matrices should use the 3D definition, and should be of size 3x3x3. The padding and cropping should also be applied in the 3 dimensions.

---
Algorithm A. Python-based pseudo-code for computing Green's function
---

```python
# Inputs:
# image_size: The size of the image
# pad: The padding to add around the image

# Find the size of the desired matrices
pad ← 4
green_function_size ← image_size + 2 * pad

# Create the Dirac and Laplace kernels
dirac ← zeros(green_function_size)
dirac[1, 1] ← 1
laplace ← zeros(green_function_size)
laplace[0:3, 0:3] ← [[0,  1,  0],
                     [1, -4,  1],
                     [0,  1,  0]]

# Compute the Green's function, and set the integration constant to 0
green_F ← fft(dirac) / fft(laplace)
green_F[0, 0] ← 0
```

---
Algorithm B. Python-based pseudo-code for solving the padded Laplacian
---

```python
# Inputs:
# padded_L: The Laplacian of the padded image
# green_F: The result of Algorithm A

# Solving the padded Laplacian using Green's function convolution
I ← ifft(fft(padded_L) * green_F)

# Integration constant and cropping
I ← I - I[0, 0]
I ← I[pad:-pad, pad:-pad]
```

## 2.3   Image classification

Due to our previous work showing the improvement of saliency results [12] and the enabling of an unlimited receptive field, one of the hypotheses for the current work is that the GF will improve many different kinds of CNN for image analysis. To explore this idea, we start with the task of image classification since it is one of the most studied problems with the most straightforward CNN architectures [9].

Hence, we started by studying the effect of the GF on a Google-net architecture applied on the MNIST [13,14] dataset. The MNIST dataset has 70,000 images of handwritten digits with a total of 10 classes (one per digit) [13,14]. It is one of the most simple classification datasets since digits are easier to classify than real images with an accuracy of 99.3% in 1998 [14]. The Google-net contains 6 consecutive inception layers and is usually used for more complex image classification [1]. For the simpler task of digits recognition, we only used 2 inception layers with 16 channels per layer in the first inception module and 32 channels per layer in the second module. The networks are coded using Tensorflow® with an Adam optimizer [8], a learning rate of $10^{-4}$ and a batch size of 50 images. These parameters were selected using a grid search to maximize the results of the standard network (without GF), but the number of parameters was kept low since the MNIST task is simple.

## 2.4 Gradient integration and derivative

To try to improve the results of the Google-net with the GF, we developed the GID (Gradient Integration Derivative). Then, we added a GID layer to each of the inception modules as shown in Figure 1. The GID layer is explained in Figure 2 and uses a Conv-layer after its input without an activation function or bias, meaning that the layer is simply a linear combination of the previous layer. It is meant to allow negative values since the input is strictly positive due to the Relu activation.

The GID is based on the idea that the features are a vector field similar to a gradient [9] or to Gabor functions which are mostly similar to Gaussian derivatives [9,15]. Hence, the GID computes the derivatives $dx$ and $dy$ of the features to obtain a Laplacian. For the integration, the GID uses the GF developed in section to solve the Laplacian. This is followed by the derivative step of the GID to compute a new vector field similar to the input one.

In addition to the GID layer, some simpler variants are also possible, such as the Laplacian integration and gradient integration, which are also shown in Figure 2. It is important to note that the GF-based layers do not need to be applied on all the input channels, they can only be applied to some of the channels.

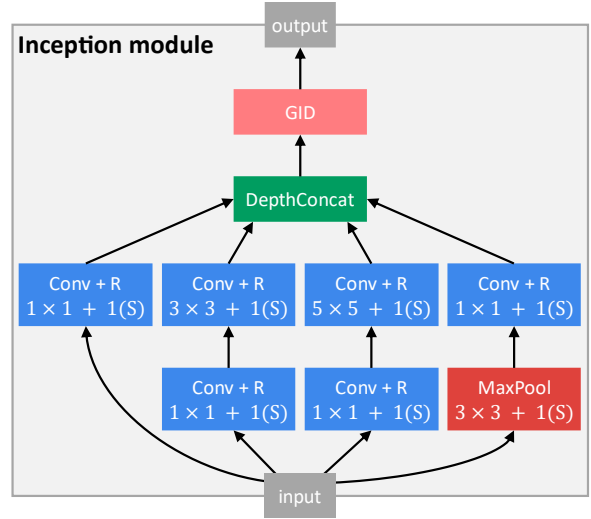

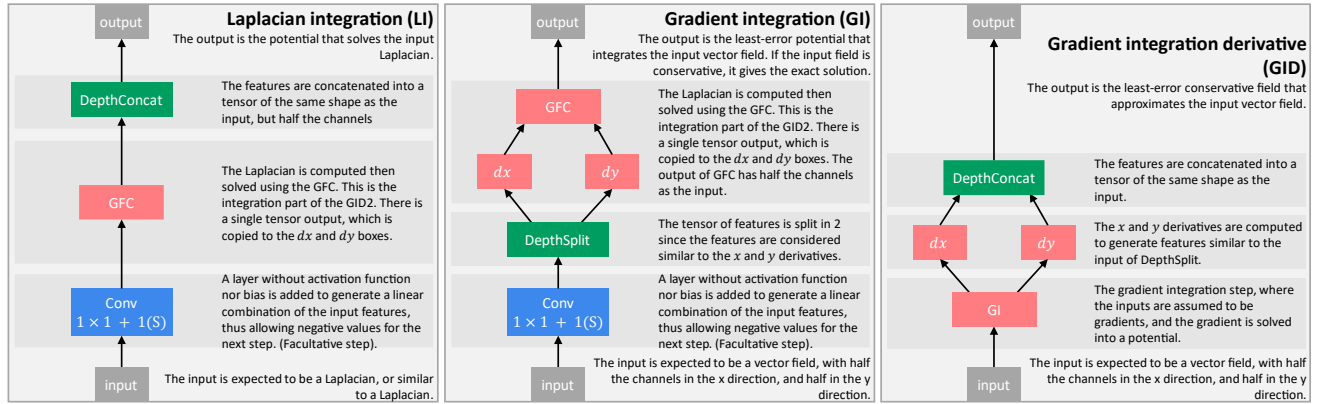

Figure 1: Inception module of the Google-net [1] with an added GDI2 or GDI3 layer. The Conv + R indicate a convolutional layer with a Relu activation function, the $n \times n$ means that the kernel size is $n$ and the $m(S)$ means that the stride is $m$.

Figure 2: Internal structure of the different GF-based layers, notably the Laplacian integration (LI), gradient integration (GI) and gradient integration derivative (GID). The $dx$ and $dy$ represent numerical derivatives and the Conv-layer does not use an activation function or bias; the $n \times n$ means that the kernel size is $n$ and the $m(S)$ means that the stride is $m$. GFC is the Green's function convolution presented in Algorithm B.

## 3 Results and discussion

### 3.1 Prototype and early results for the classification CNN

When testing the smaller Google-net composed of 2 inception modules on the MNIST dataset, we observe that the added GID significantly improves the results compared to the standard network, with all other parameters being the same. The convergence to a 97% validation success rate is around 5.1 times faster with GID. Furthermore, after 20,000 iterations, the model with the GID maxed to a smoothed validation accuracy of 98.80%, compared to 98.35% without the GID, which is a 27% decrease in the error rate. These results are observed in Figure 3 where the orange line is strictly better (higher accuracy) than the blue line, especially for a low number of iterations. Furthermore, the orange line appears smoother than the blue line, meaning that it was easier for the gradient descent to converge to a minimum when using GID. A downside of the added GID is that the training time is 2.0 times longer, but this still means that the convergence to 97% is 2.5 times faster than the standard network.

This same behavior was observed using different random seeds and using different hyperparameters. It is important to note that the GID layer adds no additional parameter, except for a single weight per feature, which cannot explain the difference in performance nor the decreased convergence time.

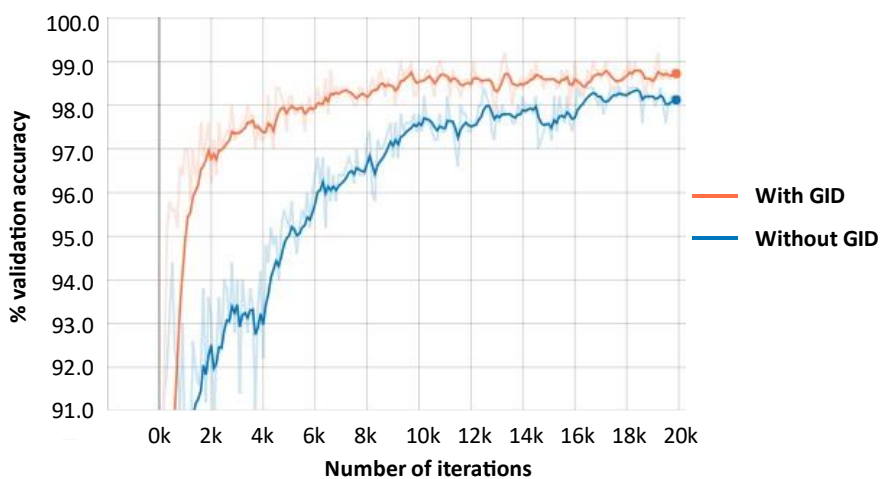

Figure 3: Validation accuracy on the MNIST dataset for a smaller Google-net with 2 inception layers. The blue line is the standard network and the orange line is the same network with the added GID in each inception layer.

Before having a final verdict on the performance of the GID, it is necessary to perform varied tests with deeper CNN and more complex images. However, we believe that GID will still be able to improve the results for many reasons. First, the integration with a Green's function followed by a derivative allows transforming the initial vector field of features into a conservative field, as demonstrated by Beaini et al. [11]. This conservative field has more continuous and smoother features with an unlimited receptive field between the given features.

We also believe that the conservative field features "make more physical sense": for example, if we take a human picture and ask to identify the head, a non-conservative field could select an open shape, such as a ¾ of a circle. This does not make physical sense since the head feature should be a closed shape. For the conservative field, such open shape will be forced to close by a smooth gradient such as demonstrated by Beaini et al. [16]. Hence, the conservative feature field means that the features are physically interpretable and can be integrated into an existing potential solution.

Moreover, we know that detecting features in every possible orientation requires at least 2 kernels with non-colinear features, the same way that we need at least 2 vectors to generate the full 2D space. In that sense, the GID is very useful since it regularizes the features by encouraging half of the features to be perpendicular to the other half to allow integration with the GF. All those reasons put together are believed to accelerate the optimization of the CNN by encouraging the gradient descent to follow a better optimization path.

In summary, we developed the GID which integrates the CNN features using a Green's function and derivates them back to give features represented by a conservative field. The GID was tested using a 2 inception layers Google-net on the MNIST handwritten digits dataset. The GID proved to reduce the final convergence error of 27%, requires 5.1 times fewer iterations to converge and has a smoother validation curve.

We are confident that this work can generalize to more tasks, due to the early results and the previous strong success of the Green's function (GF) for salient object detection [12], which allowed the network to generalize to more complex, darker and noisier images. However, more tests of the GID are required to verify if it helps improve the results with more complex networks and images.

## 3.2   Future work: Generative networks

In addition to using the GF for classification networks, we also believe the GF can be used for generative networks (GN) such as the popular generative adversarial networks (GAN) [17,18]. We expect the GF to have many advantages such as the regularization of the GN, the generation of a gradient, and the unlimited receptive field.

**Regularizing the GN.** Since the GF was shown to regularize a CNN by making it learn only features that are physically possible, we believe that it will help the GN focus on physically possible images. For example, a standard GN could generate a shape, but without properly closing its boundaries. However, using the GF, we believe that such an option will be avoided, thus improving the training of the GN.

**Generating the gradient.** Another aspect is that the GN could be used to generate the gradient of the image instead of generating the image. Then, the GF will be used to integrate the gradient into the desired image. This will mimic how humans generate images by drawing the contours of an object before filling it up. By using the GI layer from Figure 2, or the hybrid GIS layer proposed by Beaini et al. [12], the GN will be able to combine both gradient-domain and image-domain information for the generation. In fact, the field of gradient-domain image painting or editing is well developed for human software [10,11,19,20].

**Unlimited receptive field.** Finally, the GF will enable unlimited receptive field between the pixels of the generated image, which is fundamental in ensuring that the pixels are generated based on global information.

### 3.3 Limitations

Known limitations are the computational complexity of the FFT algorithm being $O(n \log n)$, with $n$ being the total number of pixels, which is greater than the linear complexity of the CNN. However, when $n$ is very large such as for a 32x32 image, the logarithmic term is negligible compared to the linear term. Another limitation is that these layers are only useful for CNN of 2D, 3D or more dimensions since the integral of a 1D vector is trivial and does not require the Green's function.

## 4 Conclusion

Building on the success of the Green's function (GF) inside CNNs for salient object detection, we generalized the usage of GF by developing integration layers that can be used in any kind of CNN. We showed an example of a classification task on a handwritten digit dataset and demonstrated that the GF-based layer significantly improved the results and the training of the network. However, more work is needed to demonstrate the advantage of GF-based layers for more complex and diverse classification tasks and different network architectures.

## References

[1] Szegedy C, Liu W, Jia Y, Sermanet P, Reed S, Anguelov D, et al. Going Deeper with Convolutions. Comput. Vis. Pattern Recognit. CVPR, 2015.

[2] He K, Zhang X, Ren S, Sun J. Deep Residual Learning for Image Recognition. ArXiv151203385 Cs 2015.

[3] Cheng MM. Richer Convolutional Features for Edge Detection. 南开大学媒体计算实验室 2017. https://mmcheng.net/rcfedge/ (accessed May 30, 2018).

[4] Xie S, Tu Z. Holistically-Nested Edge Detection n.d.:9.

[5] Hou Q, Liu J, Cheng M-M, Borji A, Torr PHS. Three Birds One Stone: A Unified Framework for Salient Object Segmentation, Edge Detection and Skeleton Extraction. ArXiv180309860 Cs 2018.

[6] Hou Q, Cheng M-M, Hu X-W, Borji A, Tu Z, Torr P. Deeply supervised salient object detection with short connections. IEEE Trans Pattern Anal Mach Intell 2018:1–1. https://doi.org/10.1109/TPAMI.2018.2815688.

[7] Li G, Yu Y. Deep Contrast Learning for Salient Object Detection, 2016, p. 478–87.

[8] Kingma DP, Ba J. Adam: A Method for Stochastic Optimization. ArXiv14126980 Cs 2014.

[9] Goodfellow I, Bengio Y, Courville A. Deep Learning. MIT Press; 2016.

[10] Tanaka M, Kamio R, Okutomi M. Seamless Image Cloning by a Closed Form Solution of a Modified Poisson Problem. SIGGRAPH Asia 2012 Posters, New York, NY, USA: ACM; 2012, p. 15:1–15:1. https://doi.org/10.1145/2407156.2407173.

[11] Beaini D, Achiche S, Nonez F, Brochu Dufour O, Leblond-Ménard C, Asaadi M, et al. Fast and Optimal Laplacian and Gradient Solver for Gradient-Domain Image Editing using Green Function Convolution 2018.

[12] Beaini D, Achiche S, Duperré A, Raison M. Deep green function convolution for improving saliency in convolutional neural networks. Vis Comput 2020. https://doi.org/10.1007/s00371-020-01795-8.

[13] MNIST handwritten digit database, Yann LeCun, Corinna Cortes and Chris Burges n.d. http://yann.lecun.com/exdb/mnist/ (accessed February 16, 2017).

[14] Lecun Y, Bottou L, Bengio Y, Haffner P. Gradient-based learning applied to document recognition. Proc IEEE 1998;86:2278–324. https://doi.org/10.1109/5.726791.

[15] Kumar A, Pang GKH. Defect detection in textured materials using Gabor filters. IEEE Trans Ind Appl 2002;38:425–40. https://doi.org/10.1109/28.993164.

[16] Beaini D, Achiche S, Nonez F, Raison M. Computing the Spatial Probability of Inclusion inside Partial Contours for Computer Vision Applications. ArXiv180601339 Cs Math 2018.

[17] Goodfellow I, Pouget-Abadie J, Mirza M, Xu B, Warde-Farley D, Ozair S, et al. Generative Adversarial Nets. In: Ghahramani Z, Welling M, Cortes C, Lawrence ND, Weinberger KQ, editors. Adv. Neural Inf. Process. Syst. 27, Curran Associates, Inc.; 2014, p. 2672–2680.

[18] Radford A, Metz L, Chintala S. Unsupervised Representation Learning with Deep Convolutional Generative Adversarial Networks. ArXiv151106434 Cs 2015.

[19] Pérez P, Gangnet M, Blake A. Poisson image editing. ACM SIGGRAPH 2003 Pap., San Diego, California: Association for Computing Machinery; 2003, p. 313–318. https://doi.org/10.1145/1201775.882269.

[20] Real-time gradient-domain painting | ACM Transactions on Graphics n.d. https://dl.acm.org/doi/abs/10.1145/1360612.1360692 (accessed March 11, 2020).